\documentclass[11pt]{article}

\usepackage[final]{acl}

\usepackage{times}
\usepackage{latexsym}

\usepackage[T1]{fontenc}

\usepackage[utf8]{inputenc}

\usepackage{microtype}

\usepackage{inconsolata}

\usepackage{graphicx}
\usepackage{booktabs}
\usepackage{amsmath}
\usepackage{algorithm}
\usepackage{algpseudocode}
\usepackage{pifont}

\usepackage{booktabs}
\usepackage{tabularx}

\usepackage{listings}
\usepackage{xcolor}
\usepackage{multirow}
\usepackage{comment}

\usepackage[table]{xcolor}

\definecolor{panelgray}{HTML}{F2F2F2}
\definecolor{oursblue}{HTML}{EAF3FF}

\usepackage{soul}

\lstdefinestyle{promptstyle}{
  basicstyle=\ttfamily\scriptsize,
  breaklines=true,
  breakatwhitespace=false,
  columns=fullflexible,
  keepspaces=true,
  frame=single,
  framerule=0.3pt,
  rulecolor=\color{black!40},
  xleftmargin=0.25em,
  xrightmargin=0.25em,
  aboveskip=0.5em,
  belowskip=0.5em
}

\title{APDMem: Agent-Controlled Progressive Disclosure \\for Query-Adaptive Long-Term Memory}

\author{Chin-Lun Fu~~~ Anagha Kulkarni~~~ Hong Ni~~~ Behrouz Madahian
  \\
  Machine Learning Center of Excellence, JPMorgan Chase \& Co.
  \\
  \small{
    \texttt{chin-lun.fu@jpmchase.com},
    \texttt{anagha.p.kulkarni@jpmorgan.com}
  }
  \\
  \small{
    \texttt{hong.ni@jpmchase.com},
    \texttt{behrouz.madahian@jpmchase.com}
  }
}

\begin{document}
\maketitle
\begin{abstract}
Personalized LLM assistants must recover sparse evidence from long conversation histories across queries of varying complexity. We introduce \textbf{APDMem} (\textbf{A}gent-controlled \textbf{P}rogressive \textbf{D}isclosure \textbf{Mem}ory), a hierarchical long-term memory architecture that applies progressive disclosure to memory retrieval. Rather than relying on a flat memory store or fixed retrieval granularity, APDMem represents conversation history as four progressively detailed layers: thematic summaries, personalized key facts, turn-level evidence notes, and raw messages. 
At inference time, a \textit{controller} applies progressive disclosure to the memory hierarchy: it first reads high-level summaries and drills into finer evidence only when needed.
This creates an adaptive cost--fidelity trade-off: simple queries can terminate early, while complex temporal, multi-hop, or exact-evidence queries trigger deeper inspection. A \textit{note synthesizer} converts retrieved evidence into a query-focused structure that consolidates facts, orders events, and flags contradictions before final answer generation. Experiments on \textsc{LongMemEval} show that APDMem achieves strong performance for long-context memory reasoning while accessing only 8\% of the total conversations.
\end{abstract}

\section{Introduction}

Large Language Model (LLM) agents have shown strong performance on complex, multi-step tasks. However, personalized agents must operate across long-term, multi-session interactions, where relevant information may be scattered across many past conversations \cite{xu2026toward, liu-etal-2025-persona, yu-etal-2026-agentic}. Such agents must remember user preferences, factual details, temporal sequences, and recurring themes over time \cite{chhikara2025mem0, NEURIPS2025_19909c36, fang2026lightmem, liu2026simplememefficientlifelongmemory, rasmussen2025zep}. Without effective memory, agents suffer from contextual amnesia: they may fail to recall prior facts, lose track of evolving goals, or overlook cross-session connections.

Prior works on memory systems improve long-term recall by transforming conversation history into specialized representations: MemoryBank uses a structured memory store for personalized dialogue \cite{zhong2024memorybank}; MemGPT manages information via virtual context and external memory \cite{packer2024memgptllmsoperatingsystems}; A-MEM builds interconnected, evolving memory networks \cite{NEURIPS2025_19909c36}; and Zep constructs temporal, entity-centric memory graphs \cite{rasmussen2025zep}. Other systems reduce cost through compression or consolidation: LightMem uses staged memory processing \cite{fang2026lightmem}, while SimpleMem retrieves over compact semantic memories \cite{liu2026simplememefficientlifelongmemory}. While these designs reduce retrieval cost, they also introduce additional memory machinery and require committing conversational context to external memory stores, graph nodes, or other abstractions.

We introduce \textbf{APDMem} (\textbf{A}gent-controlled \textbf{P}rogressive \textbf{D}isclosure \textbf{Mem}ory), a hierarchical memory system for efficient context recall in personalized agents. APDMem treats long-term memory access as a query-adaptive disclosure problem rather than a one-shot retrieval problem. Inspired by \textbf{progressive disclosure} \cite{nielsen1993usability}, it exposes coarse information first and reveals finer evidence as needed. APDMem stores conversations as lightweight Markdown layers---thematic summaries ($L_0$), personalized key facts ($L_1$), turn-level evidence notes ($L_2$), raw messages ($L_3$)---rather than converting them into a specialized database, graph, or compressed-only representation. 

Unlike existing systems that fix the retrieved representation in advance, APDMem only materializes thematic summaries at ingestion time while preserving raw messages underneath. At inference time, the \textit{Progressive Disclosure Controller} expands on demand into finer details like personalized key facts, turn-level evidence notes, or raw messages. After sufficient evidence is collected, the controller drafts an answer. For final answer generation, \textit{Organized Notes Synthesizer} structures the evidence retrieved by the controller along with the draft answer into organized notes. These notes consolidate relevant evidence, align it with the query and provide the answer generator with structured grounding context instead of raw retrieved snippets, reducing noise and improving answer accuracy.

APDMem frames long-term memory retrieval as a query-adaptive disclosure problem:
it must determine \textit{which} conversations matter and \textit{how much detail}
to expose. Concretely, APDMem introduces two coupled dimensions: (1) hierarchical memory organization (coarse summaries linked to finer-grained evidence) and (2) progressive disclosure as an explicit architectural principle governing how that hierarchy is accessed. That is, the agent is designed to open only the minimum memory level required, exposing coarse views first and revealing finer evidence only when needed, with the disclosure depth controlled through tool-mediated interactions. Our primary contributions are:
\begin{itemize}
    \item \textbf{Coarse-to-fine memory formulation.}
    We introduce a \emph{Multi-Resolution Memory Substrate} that preserves raw
    messages beneath progressively abstract memory layers.

    \item \textbf{Agent-controlled query-adaptive memory disclosure.}
    We introduce a \emph{Progressive Disclosure Controller} that decomposes
    queries, handles temporal constraints, and drills into deeper memory layers
    when needed.

    \item \textbf{Organized Notes for grounded generation.}
    We introduce an \emph{Organized Notes Synthesizer} that grounds retrieved
    evidence and the controller's draft answer into structured notes.

    \item \textbf{LongMemEval validation.}
    We show that APDMem achieves strong performance on LongMemEval
    \cite{wu2025longmemeval} while accessing only 8\% of the conversations.
\end{itemize}

\section{Related Work}

\subsection{Memory Representations for LLM Agents}

Personalized agents require persistent memory to maintain continuity across interactions \cite{zhang-etal-2026-personaagent, xu2026toward}. Prior works frame this as a long-term memory problem: agents must retain user-specific facts and temporal context across sessions to avoid contextual amnesia \cite{chhikara2025mem0, fang2026lightmem, liu2026simplememefficientlifelongmemory}. Memory systems differ mainly in how they represent past interactions. Profile- or fact-based methods store distilled user information \cite{zhong2024memorybank, chhikara2025mem0}; episodic systems retrieve prior interaction traces \cite{packer2024memgptllmsoperatingsystems, NEURIPS2025_19909c36}; graph-based systems such as Zep organize temporal entity-relation memories \cite{rasmussen2025zep}; and compression-based systems such as LightMem and SimpleMem reduce cost through staged consolidation or compact semantic memories \cite{fang2026lightmem, liu2026simplememefficientlifelongmemory}. These approaches often require graph conversion, compression, or specialized memory objects, creating trade-offs between efficiency, fidelity, and complexity. APDMem instead stores conversations as transparent Markdown layers preserving the original conversation while providing compact abstractions for retrieval.

\begin{figure*}[!ht]
\centering
\includegraphics[width=0.9\textwidth]{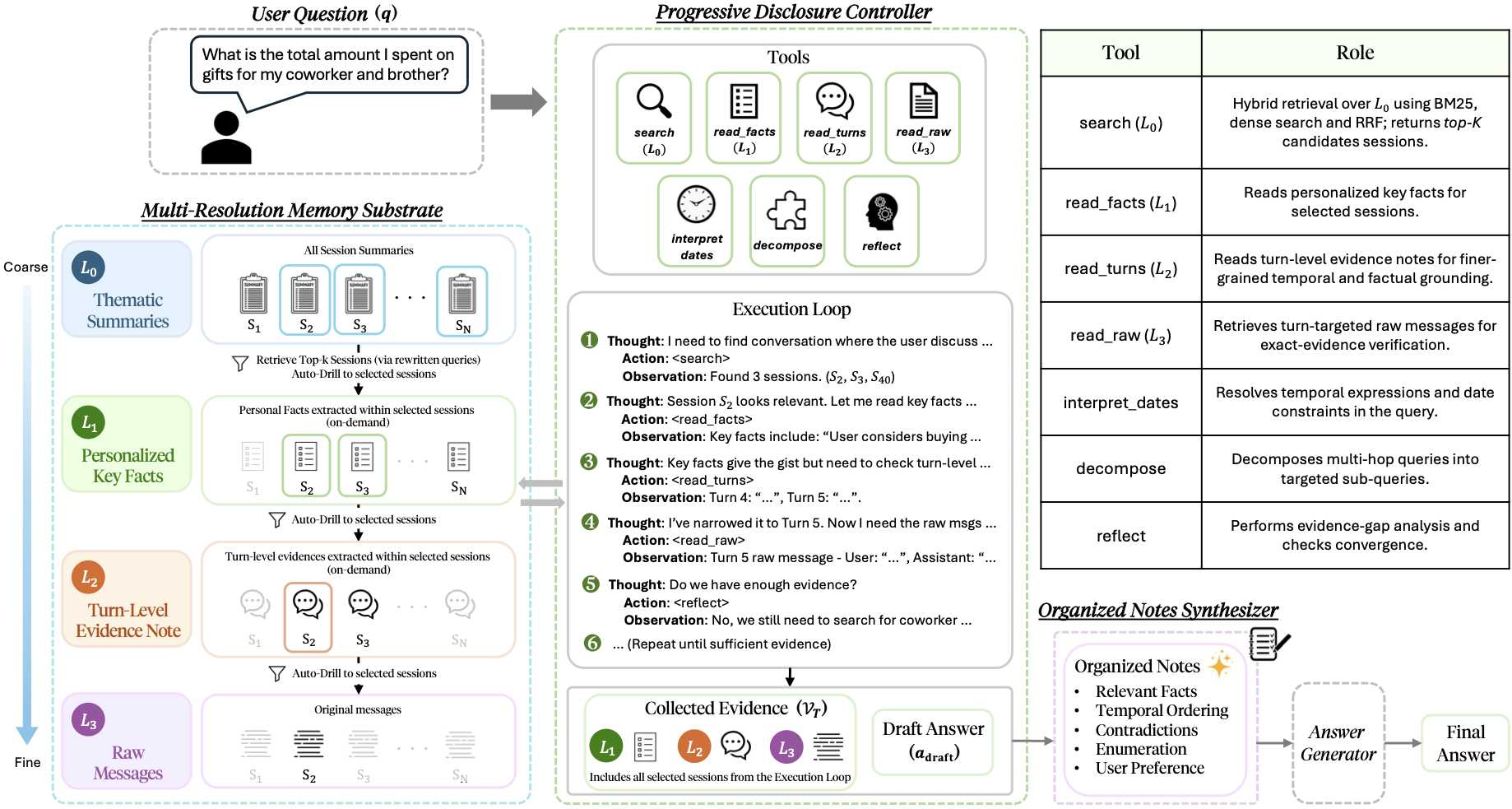}
\caption{APDMem overview. The \textit{Multi-Resolution Memory Substrate} stores conversations as layered views from thematic summaries ($L_0$) to raw messages ($L_3$). The \textit{Progressive Disclosure Controller} uses tool calls to retrieve and drill down only as needed. The \textit{Organized Notes Synthesizer} then consolidates collected evidence into structured notes for grounded answer generation. The upper-right panel summarizes tools available to the controller.}
\label{fig:architecture}
\end{figure*}

\subsection{Progressive Disclosure}

Memory systems organize information at different granularities---such as summaries, passages, chunks, or episodic traces---to improve retrieval over large contexts \cite{sarthi2024raptor, gao2023retrieval, izacard-grave-2021-leveraging, 10.1145/3397271.3401075, NEURIPS2020_6b493230, asai2024selfrag, trivedi-etal-2023-interleaving}. APDMem builds on this idea through \textit{progressive disclosure} \cite{nielsen1993usability}, exposing coarse memory views first and revealing finer evidence only when needed. To our knowledge, APDMem is the first system to make progressive disclosure an explicit architectural principle for agent memory, with an agent controlling disclosure depth through tool-mediated interactions.

\subsection{Evidence-Grounded Generation}

Complex queries involving temporal constraints, ambiguous references, user-specific details, multi-hop reasoning, or enumeration often require iterative retrieval and verification \cite{yao2023react, NEURIPS2023_d842425e, press-etal-2023-measuring, trivedi-etal-2023-interleaving}. Prior work has shown the value of organizing or reasoning over retrieved evidence before answer generation \cite{yu-etal-2024-chain, gao2023retrieval, asai2024selfrag}. Inspired by Chain-of-Note \cite{yu-etal-2024-chain}, APDMem converts retrieved evidence into organized notes that consolidate relevant facts, temporal order, contradictions, enumeration evidence, and answer hints into a structured grounding context for the answer generator.

\section{APDMem}

As shown in Figure~\ref{fig:architecture}, APDMem consists of three components: (1) a \emph{Multi-Resolution Memory Substrate} that stores each conversation across four layers of increasing abstraction (\S\ref{sec:memory}); (2) a \emph{Progressive Disclosure Controller} that searches, drills into, and reflects over the memory hierarchy through tool-mediated interactions (\S\ref{sec:agent}); and (3) an \emph{Organized Notes Synthesizer} that structures retrieved evidence for final answer generation (\S\ref{sec:organized-notes}).


\subsection{Multi-Resolution Memory Substrate}
\label{sec:memory}

Given a conversation history, we partition it into $N$ sessions $\mathcal{S}={S_1,\ldots,S_N}$, where each session $S_i$ is a contiguous conversation episode consisting of turns, and each turn is a user--assistant exchange. APDMem represents each session as a four-layer memory tuple, $S_i=(L_0,L_1,L_2,L_3)$. $L_0$ contains \textit{thematic summaries} with resolved pronouns, normalized dates, keywords, and tags, and serves as the primary indexing layer. $L_1$ contains \textit{personalized key facts}: session facts about user information, recommendations, plans, preferences, and constraints. $L_2$ contains \textit{turn-level evidence notes}: turn-level factual bullets that preserve temporal order for fine-grained evidence reasoning. $L_3$ contains \textit{raw messages}: the verbatim user--assistant messages retained for exact evidence recovery.\footnote{Construction details, prompts, and real output examples for each layer are provided in Appendix~\ref{app:memory-details}.}

\paragraph{Lazy materialization.}
Only $L_0$ and $L_3$ are materialized at ingestion time, whereas $L_1$ and $L_2$ are generated on demand when first requested by the progressive disclosure controller and then cached for reuse, reducing write-time cost.


\subsection{Progressive Disclosure Controller}
\label{sec:agent}

The \textit{Progressive Disclosure Controller} implements memory access as a
bounded, query-adaptive control process. Given a query $q$, the controller's objective is to gather sufficient evidence for answering $q$ while avoiding unnecessary disclosure of finer memory. It maintains an evidence set $\mathcal{V}_t$ at
step $t$. Each element in $\mathcal{V}$ is a retrieved view of a session $S_i$ at
abstraction level $L_\ell$ for $\ell \in \{0,1,2,3\}$.

At each step, the controller selects a tool action
$a_t \in \mathcal{A}$, where
$\mathcal{A}=\{\texttt{search}, \texttt{read\_facts}, \\ \texttt{read\_turns}, 
\texttt{read\_raw}, \texttt{interpret\_dates}, \\ \texttt{decompose},
\texttt{reflect}\}$ is the action space.  Memory-read actions correspond to
different resolutions of the \textit{Multi-Resolution Memory Substrate}: finer
layers expose higher-fidelity evidence but incur greater token and latency
cost, i.e., $c(L_0)<c(L_1)<c(L_2)<c(L_3)$ and
$f(L_0)<f(L_1)<f(L_2)<f(L_3)$, where $c(\cdot)$ denotes access cost and
$f(\cdot)$ denotes evidence fidelity. The controller can therefore be viewed as optimizing a cost--fidelity
trade-off:
\begin{equation}
  \min_{a_{1:T}} \sum_{t=1}^{T} c(a_t)
  \quad
  \text{s.t.}
  \quad
  \mathrm{sufficient}(\mathcal{V}_T, q) = 1,
  \label{eq:disclosure-objective}
\end{equation}
where $\mathrm{sufficient}(\mathcal{V}_T,q)$ indicates whether the
collected evidence is adequate for answering the query. In practice, sufficiency is
estimated through evidence reflection and bounded by the convergence criteria described in \S\ref{sec:convergence}. Unlike flat retrieval systems, APDMem selects both the relevant memory
region and the required evidence resolution.

\begin{table*}[t]
\centering
\scriptsize
\setlength{\tabcolsep}{5.2pt}
\renewcommand{\arraystretch}{1.05}
\resizebox{0.7\textwidth}{!}{
\begin{tabular}{@{}llccccccc@{}}
\toprule
\textbf{Backbone} & \textbf{Method} 
  & \textbf{Temp.} & \textbf{Multi-S.}
  & \textbf{K-Upd.} & \textbf{SS-User}
  & \textbf{SS-Asst.} & \textbf{SS-Pref.}
  & \textbf{Avg.} \\
\midrule
\multirow{6}{*}{GPT-4.1}
  & Full-context$^{\ast}$ & 51.9 & 39.1 & 70.5 & 65.7
    & \underline{96.4} & 16.7 & 56.7 \\
  & Mem0$^{\ast}$ & 43.6 & 54.9 & 75.6 & 54.3
    & 39.3 & \textbf{83.3} & 58.5 \\
  & RAPTOR$^{\dagger}$ & 38.4	& 51.9 & 68.0 & \underline{95.7}
    & 92.9 & 60.0 & 62.0 \\
  & LightMem$^{\ast}$ & 85.0 & 57.9 & \underline{89.7} & 87.1
    & 71.4 & 70.0 & 76.9 \\
  & SimpleMem$^{\ast}$ & \underline{86.5} & \textbf{81.2} & 80.8 & \textbf{98.6}
    & 76.8 & \underline{80.0} & \underline{84.0} \\
  & \cellcolor{oursblue}APDMem$^{\dagger}$ & \cellcolor{oursblue}\textbf{90.2} & \cellcolor{oursblue}\underline{77.4} & \cellcolor{oursblue}\textbf{91.0} & \cellcolor{oursblue}92.9
    & \cellcolor{oursblue}\textbf{100.0} & \cellcolor{oursblue}\underline{80.0} & \cellcolor{oursblue}\textbf{87.8} \\
\midrule
\multirow{6}{*}{GPT-4.1-mini}
  & Full-context$^{\ast}$ & 27.1 & 30.1 & 41.0 & 47.1
    & 32.1 & 60.0 & 39.6 \\
  & Mem0$^{\ast}$ & 40.6 & 50.4 & 69.2 & 87.1
    & 48.2 & 63.3 & 59.8 \\
  & RAPTOR$^{\dagger}$ & 38.4	& 55.6 & 69.2 & \underline{90.0} & \textbf{98.2} & 66.7 & 63.4 \\
  & LightMem$^{\ast}$ & \textbf{85.7} & 47.4 & \textbf{92.3} & 88.6
    & 21.4 & \textbf{76.7} & 68.7 \\
  & SimpleMem$^{\ast}$ & \underline{83.5} & \underline{60.9} & 79.5 & 85.7
    & 75.0 & \textbf{76.7} & \underline{76.9} \\
  & \cellcolor{oursblue}APDMem$^{\dagger}$ & \cellcolor{oursblue}77.4 & \cellcolor{oursblue}\textbf{66.2} & \cellcolor{oursblue}\underline{85.9} & \cellcolor{oursblue}\textbf{91.4}
    & \cellcolor{oursblue}\underline{96.4} & \cellcolor{oursblue}\textbf{76.7} & \cellcolor{oursblue}\textbf{79.8} \\
\bottomrule
\end{tabular}}
\\[2pt]
{\scriptsize $^{\dagger}$ denotes experiments run by us; $^{\ast}$ denotes prior results from \citet{liu2026simplememefficientlifelongmemory}, reported under a unified protocol that we also follow for fair comparison.}
\caption{Performance (\% accuracy) on LongMemEval-S
($n{=}500$). The highest value in each column is shown in
\textbf{bold}; the second highest is \underline{underlined}.}
\label{tab:main}
\end{table*}

\subsubsection{Controller Workflow}
\label{sec:workflow}

A typical trajectory proceeds in three phases.

\paragraph{Query preparation.}
Queries with relative dates (e.g., ``last summer'') are
normalized against the question timestamp via
\texttt{interpret\_dates}; multi-hop or compound queries are
split into focused subqueries via \texttt{decompose}.

\paragraph{Progressive disclosure.}
\texttt{search} retrieves the top-$K$ candidate sessions over
$L_0$ (hybrid lexical + dense retrieval fused with RRF;
Appendix~\ref{app:retrieval}). 
An \emph{auto-drill} step selects the most promising unexplored sessions and \texttt{read\_facts} retrieves their $L_1$ key facts.
If these facts are insufficient, the controller descends further:
\texttt{read\_turns} surfaces $L_2$ evidence in temporal order, and
\texttt{read\_raw} retrieves $L_3$ messages restricted to the most relevant
turns within selected sessions. At each step, retrieved session information is stored in evidence set $\mathcal{V}$.

\paragraph{Reflection and drafting.}
APDMem uses reflection \cite{yao2023react,NEURIPS2023_1b44b878} to avoid both premature answering and unbounded search.
\texttt{reflect} summarizes $\mathcal{V}$, identifies missing evidence, and assesses whether additional
retrieval is likely to help. If more evidence is needed, it proposes follow-up
queries and the controller re-enters the progressive disclosure phase. Once
$\mathcal{V}$ converges (\S\ref{sec:convergence}), the controller produces a
draft answer $a_{\text{draft}}$ from the final evidence set $\mathcal{V}_T$,
which is then passed to the Organized Notes Synthesizer.

\subsubsection{Convergence Control}
\label{sec:convergence}
The controller loop terminates when (1) reflection finds no new evidence,
$|\mathcal{V}_{t}| = |\mathcal{V}_{t-1}|$, or (2) step budget
$T_{\max}$ is reached. $T_{\max}$ caps the maximum number of controller iterations and therefore bounds the additional tool calls the controller can make before it must stop escalating and return the best available answer.\footnote{In all experiments, we set $T_{\max}=20$.
Appendix~\ref{app:tmax} reports the observed iteration distribution and discusses how to choose $T_{\max}$ for other applications.}

\subsection{Organized Notes Synthesizer}
\label{sec:organized-notes}

The \textit{Organized Notes Synthesizer} converts the
query $q$, evidence set $\mathcal{V}$, and controller's draft answer
$a_{\text{draft}}$ into organized notes, a compact intermediate representation. Inspired by Chain-of-Note \cite{yu-etal-2024-chain},
this step uses the draft to verify, correct, or supplement the controller's
initial reasoning. These notes organize relevant facts, temporal
ordering, contradictions, enumeration evidence for
counting queries, and an answer hint into a query-focused grounding context.

\paragraph{Answer generation.}
The answer generator receives the organized notes and
$a_{\text{draft}}$ to produce the final response. This
synthesize-then-generate design preserves the controller's retrieval
trajectory while allowing the final answer to correct errors or omissions
surfaced during evidence synthesis.

\section{Experiments}

\paragraph{Experimental Setup.}
We evaluate on LongMemEval \cite{wu2025longmemeval} (specifically the LongMemEval-S split): 500 questions across six
categories (Table~\ref{tab:main} column headers), each paired with a per-question session haystack. We compare against Full-context (no
retrieval), Mem0~\cite{chhikara2025mem0}, LightMem~\cite{fang2026lightmem},
and SimpleMem~\cite{liu2026simplememefficientlifelongmemory}. To ensure strict comparability, all
baseline numbers are taken from \citet{liu2026simplememefficientlifelongmemory}, and APDMem reuses
the identical answer-generator prompt
(Figure~\ref{fig:prompt-answer}) and LLM-as-a-judge
(GPT-4.1-mini, $temp{=}0.0$).\footnote{Following
\citet{fang2026lightmem, liu2026simplememefficientlifelongmemory}, all numbers are with decoding temperature fixed at~0.0.} APDMem and each baseline are evaluated with GPT-4.1 and
GPT-4.1-mini as backbones; for APDMem, a single backbone drives the
\textit{Progressive Disclosure Controller}, and the \textit{Organized Notes Synthesizer}.\footnote{Full implementation details and hyperparameters are provided in Appendix~\ref{app:prompts}, \ref{app:hyperparams}, and~\ref{app:retrieval}.}

\subsection{Main Experiments}
\label{sec:main-results}

Table~\ref{tab:main} presents the main results. APDMem reaches 87.8\% with GPT-4.1 and 79.8\% with
GPT-4.1-mini, ahead of the best prior baseline
(SimpleMem) by 3.8 and 2.9 points respectively.

The gains are clearest on detail-oriented queries.
For example, on SS-Asst, APDMem reaches 100.0\% with
GPT-4.1, surpassing SimpleMem (76.8\%) and Full-context
(96.4\%) while inspecting only a small slice of the
haystack (\S\ref{sec:progressive}). This highlights
the value of drilling to raw turns.

APDMem remains strong on multi-session synthesis with
GPT-4.1, scoring 77.4\%, though SimpleMem is higher at
81.2\%. This reflects a breadth-depth trade-off: flat
summary retrieval can expose more sessions per call,
while APDMem investigates top-ranked candidates in
greater detail. Notably, this is not a systematic
weakness: with GPT-4.1-mini, APDMem achieves the best
multi-session score, showing that targeted memory navigation
can also benefit multi-session reasoning.

\subsection{Inference Efficiency}
\label{sec:efficiency}

We decompose cost into \emph{offline} (one-time $L_0$
construction) and \emph{online} (per-query controller cost,
including on-demand $L_1$/$L_2$ generation). Following
\citet{fang2026lightmem}, (a) all systems use GPT-4o-mini, (b) the
shared answer generator is excluded from online cost, and (c) numbers are
per-question averages on LongMemEval ($n{=}500$). Prior work reports only baseline construction cost, so online cost is available only for APDMem; baseline online cost is marked unavailable (-) in Table~\ref{tab:efficiency}.\footnote{See Appendix~\ref{app:efficiency-models} for a per-backbone breakdown.}

\begin{table}[t]
\centering
\footnotesize
\setlength{\tabcolsep}{3.5pt}
\renewcommand{\arraystretch}{1.05}
\resizebox{0.8\columnwidth}{!}{%
\begin{tabular}{@{}l r rr rr@{}}
\toprule
& & \multicolumn{2}{c}{\textbf{Offline (Construction)}}
  & \multicolumn{2}{c}{\textbf{Online (Controller)}} \\
\cmidrule(lr){3-4} \cmidrule(lr){5-6}
\textbf{Method} & \textbf{ACC}
  & \textbf{Tok.\,(k)} & \textbf{Lat.\,(s)}
  & \textbf{Tok.\,(k)} & \textbf{Lat.\,(s)} \\
\midrule
Mem0     & 53.6 & 1{,}152.6 & 4{,}248.5
  & \multicolumn{2}{c}{--} \\
A-MEM    & 62.6 & 1{,}605.8 & 5{,}132.1
  & \multicolumn{2}{c}{--} \\
LightMem & 68.6 &      28.3 &     283.8
  & \multicolumn{2}{c}{--} \\
\midrule
\rowcolor{oursblue}
APDMem & 74.4 & 145.4 & 101.5
  & 124.3 & 37.7 \\
\bottomrule
\end{tabular}}
\caption{Efficiency on LongMemEval-S (GPT-4o-mini). \emph{Offline}: memory construction;
\emph{Online}: controller cost (cold start: no cached state for $L_1$ \& $L_2$). Baselines
from \citet{fang2026lightmem} (no
online numbers available).}
\label{tab:efficiency}
\end{table}

Table~\ref{tab:efficiency} compares APDMem with two classes of prior memory
systems. Update-heavy methods such as Mem0 and A-MEM rely on iterative memory maintenance or consolidation,
incurring million-token and multi-thousand-second construction costs; APDMem
uses roughly one-tenth of their offline token budget and a small fraction of
their latency. By contrast, LightMem reduces offline tokens through
query-agnostic pre-compression, which can discard fine-grained details. APDMem offers a practical efficiency--accuracy trade-off: it avoids
the high overhead of update-heavy systems while reducing reliance on aggressive compression, with minimal latency.

\begin{figure*}[t]
\centering
\includegraphics[width=0.8\textwidth]{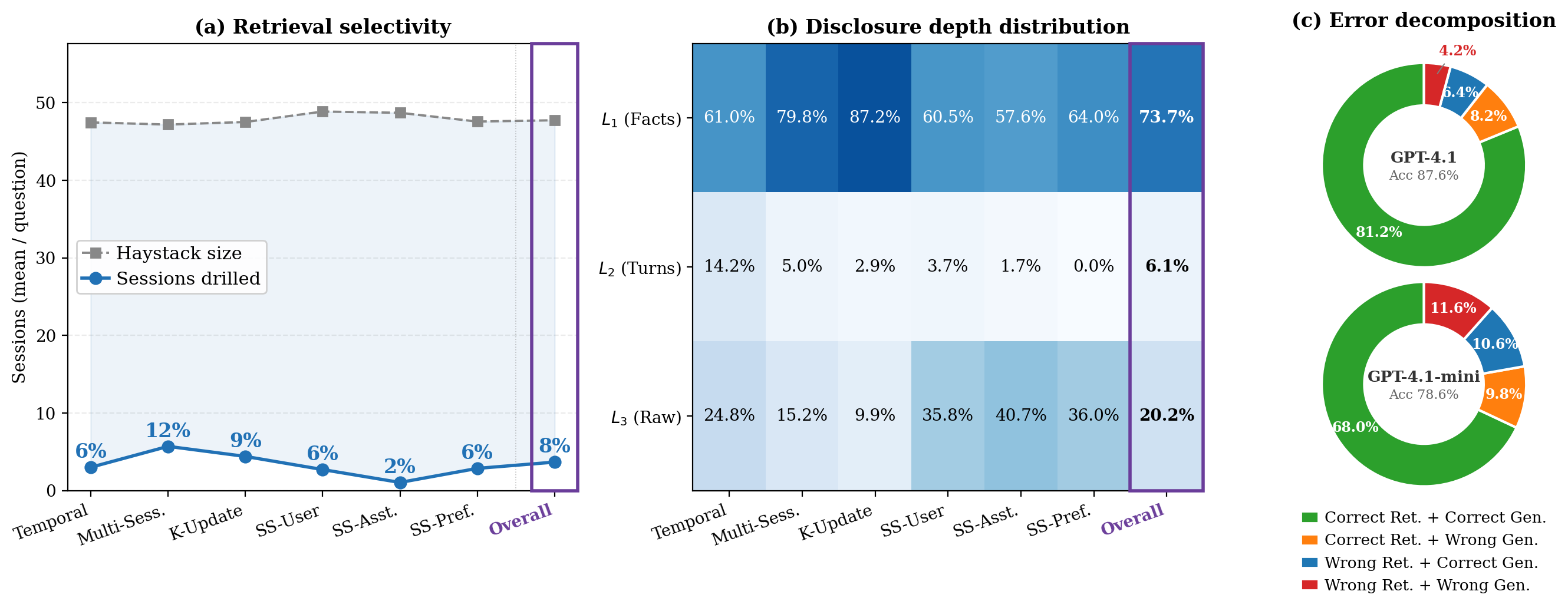}
\caption{Progressive disclosure behavior (GPT-4.1). (a)~Retrieval selectivity: fraction of haystack sessions drilled per category. (b)~Disclosure depth distribution across memory levels. (c)~Error decomposition (retrieval vs.\ generation) for GPT-4.1-mini and GPT-4.1.}
\label{fig:progressive}
\end{figure*}

\subsection{Effectiveness of Progressive Disclosure}
\label{sec:progressive}

We analyze whether progressive disclosure achieves its design
goal: locating relevant evidence with minimal memory access.
Unless stated otherwise, statistics use the GPT-4.1
configuration; the error decomposition additionally contrasts
GPT-4.1-mini with GPT-4.1 to study how backbone strength
affects retrieval versus generation failures.

\paragraph{Retrieval selectivity.}
\label{sec:behavior}
Figure~\ref{fig:progressive}(a) shows that the controller drills
into only \textbf{8\%} of the haystack on average (3.7 out
of 47.7 sessions per question). Single-session categories are
the most selective (SS-Asst.: 2\%, SS-User: 6\%), while
multi-session questions require broader access (12\%). Despite
this selectivity, APDMem achieves 87.8\% overall accuracy,
confirming that the \textit{Progressive Disclosure Controller} enables precise evidence
identification without exhaustive search.

\paragraph{Disclosure depth distribution.}
\label{sec:depth-distribution}
Figure~\ref{fig:progressive}(b) shows that the controller usually
stops shallow: among $1{,}838$ drills, 73.7\% terminate at $L_1$,
20.2\% descend to $L_3$, and 6.1\% stop at $L_2$. Depth varies by
question type: single-session queries use $L_3$ most often
(SS-Asst.\ 40.7\%, SS-User 35.8\%) because they often require exact
phrasing, whereas multi-session and knowledge-update queries remain
mostly shallow ($L_3$ rates of 15.2\% and 9.9\%). This demonstrates
that the controller adapts disclosure depth to evidence demand rather than
using a fixed retrieval granularity.

\begin{table}[t]
\centering
\small
\resizebox{0.65\columnwidth}{!}{%
\begin{tabular}{@{}llc@{}}
\toprule
\textbf{Controller} & \textbf{Generator} & \textbf{Avg.\ Acc.\ (\%)} \\
\midrule
GPT-4.1-mini  & GPT-4.1-mini & 79.8 \\
GPT-4.1       & GPT-4.1-mini & 86.8 \\
GPT-5.2       & GPT-4.1-mini & 87.8 \\
\bottomrule
\end{tabular}}
\caption{Effect of controller backbone strength with a fixed
generator on LongMemEval-S.}
\label{tab:ablation-backbone}
\end{table}

\paragraph{Error decomposition.}
\label{sec:error-analysis}
Using gold session annotations, we classify predictions by
\emph{Correct Retrieval} and \emph{Correct Generation}.
Figure~\ref{fig:progressive}(c) shows that scaling the backbone
primarily reduces retrieval errors (22.2\% $\to$ 10.6\%), while
generation errors remain near 8--10\%, making evidence discovery the
main bottleneck. This is confirmed by Table~\ref{tab:ablation-backbone}:
with the generator fixed to GPT-4.1-mini, upgrading only the controller
raises accuracy from 79.8\% to 86.8\%, and a GPT-5.2 controller reaches
87.8\%. Thus, APDMem enables heterogeneous deployment: when model budget
is limited, capacity can be allocated to the controller, where the
dominant bottleneck lies, while a lightweight model serves as the
generator.

\subsection{Ablation Studies}
\label{sec:ablation}

We ablate three aspects of APDMem on GPT-4.1 in Table~\ref{tab:ablation-tools}: the controller, memory levels, and reasoning tools. To further disentangle the effect of hierarchical memory from agentic control, we additionally include a vanilla RAG ($L_3$-only) baseline, which removes both the controller and the hierarchy and retrieves directly from raw messages. A complementary ablation of the \textit{Organized Notes Synthesizer} is reported in Appendix~\ref{app:ablation-gen}.

\begin{table}[t]
\centering
\scriptsize
\resizebox{0.7\columnwidth}{!}{
\begin{tabular}{@{}lcc@{}}
\toprule
\textbf{Configuration} & \textbf{Avg.\ Acc.\ (\%)} & \textbf{Diff} \\
\midrule
\rowcolor{oursblue}
APDMem & 87.8 & --- \\ \addlinespace[3pt]
Vanilla RAG ($L_3$-only) & 77.0 & $\downarrow$ 12.3\% \\
\midrule
\multicolumn{3}{@{}l}{\textit{Controller ablation}} \\
\cmidrule(lr){1-3}
\quad w/o \textit{Controller}$^{\dagger}$  & 83.2 & $\downarrow$ 5.2\% \\
\midrule
\multicolumn{3}{@{}l}{\textit{Memory hierarchy ablations}} \\
\cmidrule(lr){1-3}
\quad w/o \texttt{read\_facts} ($L_1$)      & 84.6 & $\downarrow$ 3.6\% \\
\quad w/o \texttt{read\_turns} ($L_2$)      & 87.4 & $\downarrow$ 0.5\% \\
\quad w/o \texttt{read\_raw} ($L_3$)        & 82.6 & $\downarrow$ 5.9\% \\
\midrule
\multicolumn{3}{@{}l}{\textit{Reasoning tool ablations}} \\
\cmidrule(lr){1-3}
\quad w/o \texttt{reflect}            & 84.6 & $\downarrow$ 3.6\% \\
\quad w/o \texttt{decompose}          & 86.0 & $\downarrow$ 2.1\% \\
\quad w/o \texttt{interpret\_dates}   & 87.6 & $\downarrow$ 0.2\% \\
\bottomrule
\end{tabular}}
\\[2pt]
{\scriptsize $^{\dagger}$RAG retrieves top-$K$ $L_0$ sessions; same sessions are populated in $L_1$--$L_3$}
\caption{Ablation on memory hierarchy levels and reasoning
tools (GPT-4.1). The ``Diff'' column is the percentage drop with respect to APDMem.}
\label{tab:ablation-tools}
\end{table}

\begin{figure*}[t]
\centering
\small
\setlength{\fboxsep}{5pt}
\begin{minipage}[t]{0.48\textwidth}
\fbox{\parbox{\dimexpr\linewidth-2\fboxsep-2\fboxrule\relax}{%
\raggedright
\textbf{Case 1: Multi-hop across sessions}\\
\textit{Q: ``How old was I when Alex was born?''}
\quad(gold: 11)\\[3pt]

\textbf{Find first fact.}
\texttt{search}($L_0$, ``Alex birth year'') $\to$ S9;\\
\texttt{read\_facts}($L_1$, S9) $\to$
\emph{``Alex is 21''}\\[2pt]

\textbf{Identify missing evidence.}
\texttt{reflect}(): need user's age.\\[2pt]

\textbf{Find second fact.}
\texttt{search}($L_0$, ``my age in 2023'') $\to$ S20;\\
\texttt{read\_facts}($L_1$, S20) $\to$
\emph{``User turned 32 in 2023-04''}\\[2pt]

\textbf{Draft and verify.}
Draft: $32-21=11$.\\
\texttt{reflect}(): verify with broader search.\\
\texttt{search}($L_0$, ``family timeline'') $\to$
S0, S16, S38, S33;\\
\texttt{read\_facts}($L_1$): no new evidence.\\[2pt]

\textbf{Converge.}
\texttt{reflect}(): sufficient $\to$ Final answer: 11\ \ding{51}\\[2pt]

\textit{Only $L_1$ evidence read; 6 sessions drilled.}
}}
\end{minipage}\hfill
\begin{minipage}[t]{0.48\textwidth}
\fbox{\parbox{\dimexpr\linewidth-2\fboxsep-2\fboxrule\relax}{%
\raggedright
\textbf{Case 2: Verbatim recall within a session}\\
\textit{Q: ``Orlando dessert shop with giant milkshakes?''}
\quad(gold: Sugar Factory at Icon Park)\\[3pt]

\textbf{Locate candidate session.}
\texttt{search}($L_0$, ``Orlando giant milkshakes'') $\to$ S10;\\
\texttt{read\_facts}($L_1$, S10) $\to$
topic \ding{51}, no shop name.\\[2pt]

\textbf{Drill for finer evidence.}
\texttt{read\_turns}($L_2$, S10) $\to$
relevant turn, still no exact name.\\[2pt]

\textbf{Recover verbatim detail.}
\texttt{read\_raw}($L_3$, S10, ``milkshakes'') $\to$\\
\hspace*{0.8em}\emph{``Sugar Factory at Icon Park''}\\[2pt]

\textbf{Converge.}
Final answer: Sugar Factory at Icon Park\ \ding{51}\\[2pt]

\textit{Descends $L_1\!\to\!L_2\!\to\!L_3$ in 1 session.}
}}
\end{minipage}
\caption{Two representative APDMem traces. Case~1 answers a cross-session multi-hop question using only $L_1$ facts, with \texttt{reflect} driving evidence-gathering and convergence. Case~2 answers a within-session verbatim question by descending the hierarchy from $L_1$ to $L_3$. The two cases highlight the breadth-vs.-depth behavior enabled by the controller.}
\label{fig:cases}
\end{figure*}

\paragraph{Results.}
\label{sec:ablation-tools}

Three trends emerge, with drops reported relative to APDMem. 
(i)~\textit{Controller and hierarchy}: replacing the controller with a non-agentic RAG baseline while preserving the $L_0$--$L_3$ hierarchy reduces accuracy from 87.8\% to 83.2\%, corresponding to a 5.2\% relative drop. This shows that query-adaptive drilling and reflection add value over static retrieval. The vanilla RAG ($L_3$-only) baseline further removes the hierarchy and retrieves only from raw messages ($L_3$), reducing accuracy to 77.0\%, corresponding to a larger 12.3\% relative drop. This comparison separates the contribution of the two components: the controller improves over hierarchical RAG, while the hierarchy itself provides a substantial gain over raw-message retrieval.
(ii)~\textit{Memory levels}: key facts
($L_1$, $\downarrow$3.6\%) and raw messages
($L_3$, $\downarrow$5.9\%) cause the largest accuracy drops,
consistent with APDMem's broad-shallow and narrow-deep retrieval
patterns. Turn-level evidence notes ($L_2$, $\downarrow$0.5\%) have a
smaller aggregate effect, serving primarily as an intermediate evidence
layer. (iii)~\textit{Reasoning tools}: reflection contributes most
($\downarrow$3.6\%), followed by decomposition
($\downarrow$2.1\%); date interpretation has little
effect ($\downarrow$0.2\%), likely because the
backbone model already handles many temporal cues.

\subsection{Case Studies}
\label{sec:cases}

Figure~\ref{fig:cases} shows two representative LongMemEval
traces illustrating how APDMem performs progressive disclosure in different scenarios.

\paragraph{Case 1.}
The first retrieved session provides Alex's age but not the
user's age, so \texttt{reflect} identifies the missing
evidence and triggers a second search. After deriving the
draft answer, a broader $L_1$ check over four additional
sessions finds no new evidence, allowing the controller to
converge without descending to $L_2$ or $L_3$.

\paragraph{Case 2.}
The controller first localizes the relevant Orlando session,
but $L_1$ only confirms the topic and does not preserve the
specific shop name. APDMem therefore drills progressively:
$L_2$ narrows the evidence region, and $L_3$ recovers the
verbatim detail needed for the final answer.

\noindent These traces illustrate APDMem's adaptive behavior:
the controller expands across sessions when evidence is
scattered, but descends within a session when the answer
depends on fine-grained details.

\section{Conclusion}

We present \textbf{APDMem}, a lightweight hierarchical memory architecture that frames long-term memory access as a query-adaptive disclosure problem. A controller first reads coarse abstractions and drills into finer-grained evidence only when needed, while a notes synthesizer organizes the surfaced evidence for grounded answer generation. On LongMemEval, APDMem achieves a substantial improvement over prior work, reaching 87.8\% accuracy with GPT-4.1. Our key findings are: (i) disclosure depth adapts to question type, with most questions answered from the high-level key-fact layer, while certain detail-oriented question categories benefit from raw-message access; (ii) controller ablations show that agent-controlled progressive disclosure is a significant contributor to performance; and (iii) APDMem offers a practical efficiency--accuracy trade-off, balancing token cost and latency while maintaining strong answer accuracy.

\section*{Limitations}

APDMem's retrieval recall depends on the quality of the $L_0$ thematic
summaries: if a summary omits a relevant detail, the corresponding turn may
receive lower BM25 and dense-retrieval scores. The keyword-field boost (Appendix
\ref{app:retrieval}) and the controller's \texttt{reflect} tool provide
partial recovery mechanisms, but initial ranking remains sensitive to summary
fidelity. To this end, developing a systematic summary-quality evaluation and monitoring for $L_0$ is a natural and important direction for future work. As one potential mitigation, summary quality can be assessed using an LLM-as-a-judge style evaluation (e.g., checking faithfulness and coverage against the underlying raw messages).

Lazy materialization of $L_1$ and $L_2$ may introduce cold-start
latency when a session is accessed for the first time. To mitigate first-access latency, $L_1$ and $L_2$ can be proactively materialized (e.g., offline or in the background). That said, our design intentionally follows progressive disclosure and lazy materialization: precomputing everything can be unnecessarily costly when many sessions (and fine-grained layers) are never queried. In production, this becomes an explicit cost–latency trade-off: deployments that prioritize low tail-latency can pre-warm likely-to-be-accessed sessions, while cost-sensitive settings can keep on-demand generation to minimize wasted computation.

Finally, APDMem does not explicitly model cross-session relationships in the way graph-based memory structures can. Exploring lightweight mechanisms
for cross-session linking is a promising direction for future work.

\section*{Disclaimer}
This paper was prepared for informational purposes in part by the Machine Learning Center of Excellence group of JPMorgan Chase \& Co. and its affiliates (``JP Morgan'') and is not a product of the Research Department of JP Morgan. JP Morgan makes no representation and warranty whatsoever and disclaims all liability, for the completeness, accuracy or reliability of the information contained herein. This document is not intended as investment research or investment advice, or a recommendation, offer or solicitation for the purchase or sale of any security, financial instrument, financial product or service, or to be used in any way for evaluating the merits of participating in any transaction, and shall not constitute a solicitation under any jurisdiction or to any person, if such solicitation under such jurisdiction or to such person would be unlawful.

\bibliography{custom}

@inproceedings{cormack2009reciprocal,
  title={Reciprocal rank fusion outperforms condorcet and individual rank learning methods},
  author={Cormack, Gordon V and Clarke, Charles LA and Buettcher, Stefan},
  booktitle={Proceedings of the 32nd international ACM SIGIR conference on Research and development in information retrieval},
  pages={758--759},
  year={2009}
}

@inproceedings{
yao2023react,
title={ReAct: Synergizing Reasoning and Acting in Language Models},
author={Shunyu Yao and Jeffrey Zhao and Dian Yu and Nan Du and Izhak Shafran and Karthik R Narasimhan and Yuan Cao},
booktitle={The Eleventh International Conference on Learning Representations },
year={2023},
url={https://openreview.net/forum?id=WE_vluYUL-X}
}

@inproceedings{press-etal-2023-measuring,
    title = "Measuring and Narrowing the Compositionality Gap in Language Models",
    author = "Press, Ofir  and
      Zhang, Muru  and
      Min, Sewon  and
      Schmidt, Ludwig  and
      Smith, Noah  and
      Lewis, Mike",
    editor = "Bouamor, Houda  and
      Pino, Juan  and
      Bali, Kalika",
    booktitle = "Findings of the Association for Computational Linguistics: EMNLP 2023",
    month = dec,
    year = "2023",
    address = "Singapore",
    publisher = "Association for Computational Linguistics",
    url = "https://aclanthology.org/2023.findings-emnlp.378/",
    doi = "10.18653/v1/2023.findings-emnlp.378",
    pages = "5687--5711"
}

@inproceedings{yu-etal-2024-chain,
    title = "Chain-of-Note: Enhancing Robustness in Retrieval-Augmented Language Models",
    author = "Yu, Wenhao  and
      Zhang, Hongming  and
      Pan, Xiaoman  and
      Cao, Peixin  and
      Ma, Kaixin  and
      Li, Jian  and
      Wang, Hongwei  and
      Yu, Dong",
    editor = "Al-Onaizan, Yaser  and
      Bansal, Mohit  and
      Chen, Yun-Nung",
    booktitle = "Proceedings of the 2024 Conference on Empirical Methods in Natural Language Processing",
    month = nov,
    year = "2024",
    address = "Miami, Florida, USA",
    publisher = "Association for Computational Linguistics",
    url = "https://aclanthology.org/2024.emnlp-main.813/",
    doi = "10.18653/v1/2024.emnlp-main.813",
    pages = "14672--14685"
}

@inproceedings{zhang-etal-2026-personaagent,
    title = "{P}ersona{A}gent: Bridging Memory and Action for Personalized {LLM} Agents",
    author = "Zhang, Weizhi  and
      Zhang, Xinyang  and
      Zhang, Chenwei  and
      Yang, Liangwei  and
      Shang, Jingbo  and
      Wei, Zhepei  and
      Zou, Henry Peng  and
      Huang, Zijie  and
      Wang, Zhengyang  and
      Gao, Yifan  and
      Pan, Xiaoman  and
      Xiong, Lian  and
      Liu, Jingguo  and
      Yu, Philip S.  and
      Li, Xian",
    editor = "Liakata, Maria  and
      Moreira, Viviane P.  and
      Zhang, Jiajun  and
      Jurgens, David",
    booktitle = "Findings of the {A}ssociation for {C}omputational {L}inguistics: {ACL} 2026",
    month = jul,
    year = "2026",
    address = "San Diego, California, United States",
    publisher = "Association for Computational Linguistics",
    url = "https://aclanthology.org/2026.findings-acl.1315/",
    doi = "10.18653/v1/2026.findings-acl.1315",
    pages = "26421--26439",
    ISBN = "979-8-89176-395-1"
}

@inproceedings{NEURIPS2023_d842425e,
 author = {Schick, Timo and Dwivedi-Yu, Jane and Dessi, Roberto and Raileanu, Roberta and Lomeli, Maria and Hambro, Eric and Zettlemoyer, Luke and Cancedda, Nicola and Scialom, Thomas},
 booktitle = {Advances in Neural Information Processing Systems},
 doi = {10.52202/075280-2997},
 editor = {A. Oh and T. Naumann and A. Globerson and K. Saenko and M. Hardt and S. Levine},
 pages = {68539--68551},
 publisher = {Curran Associates, Inc.},
 title = {Toolformer: Language Models Can Teach Themselves to Use Tools},
 url = {https://proceedings.neurips.cc/paper_files/paper/2023/file/d842425e4bf79ba039352da0f658a906-Paper-Conference.pdf},
 volume = {36},
 year = {2023}
}

@inproceedings{NEURIPS2020_6b493230,
 author = {Lewis, Patrick and Perez, Ethan and Piktus, Aleksandra and Petroni, Fabio and Karpukhin, Vladimir and Goyal, Naman and K\"{u}ttler, Heinrich and Lewis, Mike and Yih, Wen-tau and Rockt\"{a}schel, Tim and Riedel, Sebastian and Kiela, Douwe},
 booktitle = {Advances in Neural Information Processing Systems},
 editor = {H. Larochelle and M. Ranzato and R. Hadsell and M.F. Balcan and H. Lin},
 pages = {9459--9474},
 publisher = {Curran Associates, Inc.},
 title = {Retrieval-Augmented Generation for Knowledge-Intensive NLP Tasks},
 url = {https://proceedings.neurips.cc/paper_files/paper/2020/file/6b493230205f780e1bc26945df7481e5-Paper.pdf},
 volume = {33},
 year = {2020}
}

@inproceedings{
asai2024selfrag,
title={Self-{RAG}: Learning to Retrieve, Generate, and Critique through Self-Reflection},
author={Akari Asai and Zeqiu Wu and Yizhong Wang and Avirup Sil and Hannaneh Hajishirzi},
booktitle={The Twelfth International Conference on Learning Representations},
year={2024},
url={https://openreview.net/forum?id=hSyW5go0v8}
}

@inproceedings{trivedi-etal-2023-interleaving,
    title = "Interleaving Retrieval with Chain-of-Thought Reasoning for Knowledge-Intensive Multi-Step Questions",
    author = "Trivedi, Harsh  and
      Balasubramanian, Niranjan  and
      Khot, Tushar  and
      Sabharwal, Ashish",
    editor = "Rogers, Anna  and
      Boyd-Graber, Jordan  and
      Okazaki, Naoaki",
    booktitle = "Proceedings of the 61st Annual Meeting of the Association for Computational Linguistics (Volume 1: Long Papers)",
    month = jul,
    year = "2023",
    address = "Toronto, Canada",
    publisher = "Association for Computational Linguistics",
    url = "https://aclanthology.org/2023.acl-long.557/",
    doi = "10.18653/v1/2023.acl-long.557",
    pages = "10014--10037"
}

@misc{packer2024memgptllmsoperatingsystems,
      title={MemGPT: Towards LLMs as Operating Systems}, 
      author={Charles Packer and Sarah Wooders and Kevin Lin and Vivian Fang and Shishir G. Patil and Ion Stoica and Joseph E. Gonzalez},
      year={2024},
      eprint={2310.08560},
      archivePrefix={arXiv},
      primaryClass={cs.AI},
      url={https://arxiv.org/abs/2310.08560}, 
}

@inproceedings{zhong2024memorybank,
  title={Memorybank: Enhancing large language models with long-term memory},
  author={Zhong, Wanjun and Guo, Lianghong and Gao, Qiqi and Ye, He and Wang, Yanlin},
  booktitle={Proceedings of the AAAI conference on artificial intelligence},
  volume={38},
  number={17},
  pages={19724--19731},
  year={2024},
doi={https://doi.org/10.1609/aaai.v38i17.29946},
}

@inproceedings{NEURIPS2023_1b44b878,
 author = {Shinn, Noah and Cassano, Federico and Gopinath, Ashwin and Narasimhan, Karthik and Yao, Shunyu},
 booktitle = {Advances in Neural Information Processing Systems},
 doi = {10.52202/075280-0377},
 editor = {A. Oh and T. Naumann and A. Globerson and K. Saenko and M. Hardt and S. Levine},
 pages = {8634--8652},
 publisher = {Curran Associates, Inc.},
 title = {Reflexion: language agents with verbal reinforcement learning},
 url = {https://proceedings.neurips.cc/paper_files/paper/2023/file/1b44b878bb782e6954cd888628510e90-Paper-Conference.pdf},
 volume = {36},
 year = {2023}
}

@article{gao2023retrieval,
  title={Retrieval-augmented generation for large language models: A survey},
  author={Gao, Yunfan and Xiong, Yun and Gao, Xinyu and Jia, Kangxiang and Pan, Jinliu and Bi, Yuxi and Dai, Yi and Sun, Jiawei and Wang, Meng and Wang, Haofen},
  journal={arXiv preprint arXiv:2312.10997},
  url={https://arxiv.org/abs/2312.10997}, 
  year={2023}
}

@inproceedings{10.1145/3397271.3401075,
author = {Khattab, Omar and Zaharia, Matei},
title = {ColBERT: Efficient and Effective Passage Search via Contextualized Late Interaction over BERT},
year = {2020},
isbn = {9781450380164},
publisher = {Association for Computing Machinery},
address = {New York, NY, USA},
url = {https://doi.org/10.1145/3397271.3401075},
doi = {10.1145/3397271.3401075},
booktitle = {Proceedings of the 43rd International ACM SIGIR Conference on Research and Development in Information Retrieval},
pages = {39–48},
numpages = {10},
location = {Virtual Event, China},
series = {SIGIR '20}
}

@inproceedings{izacard-grave-2021-leveraging,
    title = "Leveraging Passage Retrieval with Generative Models for Open Domain Question Answering",
    author = "Izacard, Gautier  and
      Grave, Edouard",
    editor = "Merlo, Paola  and
      Tiedemann, Jorg  and
      Tsarfaty, Reut",
    booktitle = "Proceedings of the 16th Conference of the European Chapter of the Association for Computational Linguistics: Main Volume",
    month = apr,
    year = "2021",
    address = "Online",
    publisher = "Association for Computational Linguistics",
    url = "https://aclanthology.org/2021.eacl-main.74/",
    doi = "10.18653/v1/2021.eacl-main.74",
    pages = "874--880"
}

@inproceedings{
wu2025longmemeval,
title={LongMemEval: Benchmarking Chat Assistants on Long-Term Interactive Memory},
author={Di Wu and Hongwei Wang and Wenhao Yu and Yuwei Zhang and Kai-Wei Chang and Dong Yu},
booktitle={The Thirteenth International Conference on Learning Representations},
year={2025},
url={https://openreview.net/forum?id=pZiyCaVuti}
}

@article{chhikara2025mem0,
  title={Mem0: Building production-ready ai agents with scalable long-term memory},
  author={Chhikara, Prateek and Khant, Dev and Aryan, Saket and Singh, Taranjeet and Yadav, Deshraj},
  journal={arXiv preprint arXiv:2504.19413},
  url={https://arxiv.org/abs/2504.19413}, 
  year={2025}
}

@inproceedings{
sarthi2024raptor,
title={{RAPTOR}: Recursive Abstractive Processing for Tree-Organized Retrieval},
author={Parth Sarthi and Salman Abdullah and Aditi Tuli and Shubh Khanna and Anna Goldie and Christopher D Manning},
booktitle={The Twelfth International Conference on Learning Representations},
year={2024},
url={https://openreview.net/forum?id=GN921JHCRw}
}

@inproceedings{NEURIPS2025_19909c36,
 author = {Xu, Wujiang and Liang, Zujie and Mei, Kai and Gao, Hang and Tan, Juntao and Zhang, Yongfeng},
 booktitle = {Advances in Neural Information Processing Systems},
 doi = {10.52202/085713-0593},
 editor = {D. Belgrave and C. Zhang and H. Lin and R. Pascanu and P. Koniusz and M. Ghassemi and N. Chen},
 pages = {17577--17604},
 publisher = {Curran Associates, Inc.},
 title = {A-Mem: Agentic Memory for LLM Agents},
 url = {https://proceedings.neurips.cc/paper_files/paper/2025/file/19909c36f51abc4856b4560aff3d36d6-Paper-Conference.pdf},
 volume = {38, Main Conference},
 year = {2025}
}

@inproceedings{yu-etal-2026-agentic,
    title = "Agentic Memory: Learning Unified Long-Term and Short-Term Memory Management for Large Language Model Agents",
    author = "Yu, Yi  and
      Yao, Liuyi  and
      Xie, Yuexiang  and
      Tan, Qingquan  and
      Feng, Jiaqi  and
      Li, Yaliang  and
      Wu, Libing",
    editor = "Liakata, Maria  and
      Moreira, Viviane P.  and
      Zhang, Jiajun  and
      Jurgens, David",
    booktitle = "Proceedings of the 64th Annual Meeting of the {A}ssociation for {C}omputational {L}inguistics (Volume 1: Long Papers)",
    month = jul,
    year = "2026",
    address = "San Diego, California, United States",
    publisher = "Association for Computational Linguistics",
    url = "https://aclanthology.org/2026.acl-long.981/",
    doi = "10.18653/v1/2026.acl-long.981",
    pages = "21457--21483",
    ISBN = "979-8-89176-390-6"
}

@article{xu2026toward,
  title={Toward personalized llm-powered agents: Foundations, evaluation, and future directions},
  author={Xu, Yue and Chen, Qian and Ma, Zizhan and Liu, Dongrui and Wang, Wenxuan and Wang, Xiting and Xiong, Li and Wang, Wenjie},
  journal={arXiv preprint arXiv:2602.22680},
  url={https://arxiv.org/abs/2602.22680}, 
  year={2026}
}

@inproceedings{liu-etal-2025-persona,
    title = "A Persona-Aware {LLM}-Enhanced Framework for Multi-Session Personalized Dialogue Generation",
    author = "Liu, Dongshuo  and
      Wu, Zhijing  and
      Song, Dandan  and
      Huang, Heyan",
    editor = "Che, Wanxiang  and
      Nabende, Joyce  and
      Shutova, Ekaterina  and
      Pilehvar, Mohammad Taher",
    booktitle = "Findings of the Association for Computational Linguistics: ACL 2025",
    month = jul,
    year = "2025",
    address = "Vienna, Austria",
    publisher = "Association for Computational Linguistics",
    url = "https://aclanthology.org/2025.findings-acl.5/",
    doi = "10.18653/v1/2025.findings-acl.5",
    pages = "103--123",
    ISBN = "979-8-89176-256-5"
}

@inproceedings{
fang2026lightmem,
title={LightMem: Lightweight and Efficient Memory-Augmented Generation},
author={Jizhan Fang and Xinle Deng and Haoming Xu and Ziyan Jiang and Yuqi Tang and Ziwen Xu and Shumin Deng and Yunzhi Yao and Mengru Wang and Shuofei Qiao and Huajun Chen and Ningyu Zhang},
booktitle={The Fourteenth International Conference on Learning Representations},
year={2026},
url={https://openreview.net/forum?id=dyJ0GWpjJB}
}

@misc{liu2026simplememefficientlifelongmemory,
      title={SimpleMem: Efficient Lifelong Memory for LLM Agents}, 
      author={Jiaqi Liu and Yaofeng Su and Peng Xia and Siwei Han and Zeyu Zheng and Cihang Xie and Mingyu Ding and Huaxiu Yao},
      year={2026},
      eprint={2601.02553},
      archivePrefix={arXiv},
      primaryClass={cs.AI},
      url={https://arxiv.org/abs/2601.02553}, 
}

@article{rasmussen2025zep,
  title={Zep: a temporal knowledge graph architecture for agent memory},
  author={Rasmussen, Preston and Paliychuk, Pavlo and Beauvais, Travis and Ryan, Jack and Chalef, Daniel},
  journal={arXiv preprint arXiv:2501.13956},
  url={https://arxiv.org/abs/2501.13956}, 
  year={2025}
}

@book{nielsen1993usability,
  author    = {Jakob Nielsen},
  title     = {Usability Engineering},
  year      = {1993},
  publisher = {Morgan Kaufmann},
  address   = {San Francisco, CA},
  isbn      = {978-0125184069}
}

\appendix

\section{Progressive Disclosure Controller Algorithm}
\label{sec:algo}

Algorithm~\ref{alg:loop} summarizes the iterative protocol of progressive disclosure controller.

\section{System Prompts}
\label{app:prompts}

This appendix provides all system prompts used in APDMem.
Figures~\ref{fig:prompt-level0}--\ref{fig:prompt-level2} show
the memory construction prompts for $L_0$ - $L_2$;
Figures~\ref{fig:prompt-agent}--\ref{fig:prompt-answer} show
the controller, evidence reflection, organized notes
synthesis, and answer generation prompts.

\paragraph{Memory construction prompts.}
All three construction prompts (Figures~\ref{fig:prompt-level0}--\ref{fig:prompt-level2}) enforce semantic disambiguation: pronouns are resolved to named entities and relative dates are converted to absolute dates using the session timestamp as anchor.

\begin{algorithm}[t]
\caption{Progressive Disclosure Controller}
\label{alg:loop}
\small
\begin{algorithmic}[1]
\Require Query $q$, timestamp $d$, memory $\mathcal{S}$, max steps $T_{\max}$
\State $\mathcal{V}_0 \leftarrow \emptyset$;\; nudges $\leftarrow 0$
\For{$t = 1$ \textbf{to} $T_{\max}$}
  \State response $\leftarrow$ \Call{LLM}{q,d,$\mathcal{V}_{t-1}$,$\mathcal{A}$}
  \If{response contains tool calls}
    \State $\mathcal{V}_t \leftarrow \mathcal{V}_{t-1}$
    \ForAll{tool call $\tau$ in response}
      \State $\mathcal{V}_t \leftarrow \mathcal{V}_t \cup \Call{Execute}{\tau}$
    \EndFor
  \ElsIf{$|\mathcal{V}_{t-1}|$ grew since last reflection \textbf{and} nudges $< 2$}
    \State Inject nudge message;\; nudges $\leftarrow$ nudges $+ 1$
    \State $\mathcal{V}_t \leftarrow \mathcal{V}_{t-1}$
  \Else
    \State $a_{\text{draft}} \leftarrow$ response
    \State \Return $a_{\text{draft}},\mathcal{V}_{t-1}$
  \EndIf
\EndFor
\State \Return $a_{\text{draft}},\mathcal{V}_{T_{\max}}$
\end{algorithmic}
\end{algorithm}

\begin{figure}[t]
\centering
\begin{minipage}{\columnwidth}
\begin{lstlisting}[style=promptstyle]
You are a memory-indexing assistant. For each
user<->assistant turn in the conversation,
produce:
1. A 1-sentence summary (max 30 words).
2. keywords (5-8): specific names, items, dates,
   locations, actions mentioned.
3. tags (3-5): broader category labels.

Rules:
- NO pronouns -- use actual names/entities.
- NO relative dates -- convert to absolute dates
  using the [Session date] anchor.
- Each summary must be self-contained.

Session date: {session_date}

[conversation window]

Return JSON:
{"turns": [{"turn_number": 1,
  "summary": "...",
  "keywords": ["kw1", ...],
  "tags": ["tag1", ...]}, ...]}
\end{lstlisting}
\end{minipage}
\caption{Prompt for $L_0$ turn summary and keyword
generation (\S\ref{sec:memory}).}
\label{fig:prompt-level0}
\end{figure}

\begin{figure}[t]
\centering
\begin{minipage}{\columnwidth}
\begin{lstlisting}[style=promptstyle]
You are a memory-management assistant. Given a
conversation between a user and an assistant,
extract a bullet list of key facts, events,
preferences, and knowledge mentioned.

Semantic Disambiguation Rules (CRITICAL):
1. PROHIBIT all pronouns (he, she, it, they).
   Always use the actual person/item name.
2. PROHIBIT relative time expressions.
   Convert ALL time references to absolute
   dates using the [Session date] anchor.
3. Each bullet must be a complete,
   self-contained, lossless restatement
   understandable without the conversation.
4. Include specific: dates, person names, item
   names, locations, quantities, preferences.
5. Do NOT omit any factual information.

Return ONLY the bullet list (use "- " prefix).
Aim for 5-25 bullets depending on richness.
\end{lstlisting}
\end{minipage}
\caption{Prompt for $L_1$ key fact extraction
(\S\ref{sec:memory}).}
\label{fig:prompt-level1}
\end{figure}

\begin{figure}[t]
\centering
\begin{minipage}{\columnwidth}
\begin{lstlisting}[style=promptstyle]
You are a memory-management assistant. Given a
conversation between a user and an assistant,
produce per-turn atomic memory entries -- multiple
self-contained fact bullets for each user<->
assistant exchange (turn).

Semantic Disambiguation Rules (CRITICAL):
1. PROHIBIT all pronouns (he, she, it, they).
   Always use the actual person/item name.
2. PROHIBIT relative time expressions.
   Convert ALL time references to absolute
   dates using the [Session date] anchor.
3. Each bullet must be a complete, independent,
   lossless restatement -- a single atomic fact
   fully understandable without other context.
4. Include specific: dates, person names, item
   names, locations, quantities, preferences.
5. Do NOT merge multiple facts into one bullet.

Output format:

Turn 1:
- <atomic fact 1>
- <atomic fact 2>

Turn 2:
- <atomic fact 1>
- <atomic fact 2>
\end{lstlisting}
\end{minipage}
\caption{Prompt for $L_2$ turn-level evidence generation
(\S\ref{sec:memory}).}
\label{fig:prompt-level2}
\end{figure}

\paragraph{Retrieval and reasoning prompts.}
The controller (Figure~\ref{fig:prompt-agent}) operates the progressive disclosure loop, selecting tools based on question type and evidence sufficiency. The reflection prompt (Figure~\ref{fig:prompt-reflect}) performs structured gap analysis and generates follow-up queries when evidence is insufficient.

\begin{figure}[t]
\centering
\begin{minipage}{\columnwidth}
\begin{lstlisting}[style=promptstyle]
You are a personal-memory retrieval agent.
The user previously had many conversations with
an AI assistant. Those conversations have been
stored in a multi-level memory system.

Memory levels (progressive disclosure):
 Level 0 -- session summaries (per-turn
   one-liners + keywords)
 Level 1 -- key facts (bullet list extracted
   from a session)
 Level 2 -- turn summaries (self-contained,
   no pronouns, absolute dates)
 Level 3 -- raw messages (actual user<->
   assistant text)

STRATEGY GUIDELINES:
1. If the question contains relative time
   references, call interpret_dates FIRST.
2. If the question is complex, call
   decompose.
3. Call search with 3-5 diverse queries.
4. READ key facts for ALL possibly-relevant
   sessions. For counting questions, cast a
   WIDE net.
5. For counting/aggregation questions: MUST call
   reflect to check for gaps.
6. Use temporal reasoning: recent sessions'
   items likely qualify.
7. Deduplicate items across sessions.

CRITICAL: For counting questions, NEVER give
a final answer without calling
reflect at least once.
\end{lstlisting}
\end{minipage}
\caption{System prompt for the Progressive Disclosure
Controller (\S\ref{sec:agent}).}
\label{fig:prompt-agent}
\end{figure}

\begin{figure}[t]
\centering
\begin{minipage}{\columnwidth}
\begin{lstlisting}[style=promptstyle]
You are a reflection assistant for a memory
retrieval system that uses BM25 keyword matching
with English stemming and embedding similarity.

The user asked (on {question_date}): {question}

Sessions searched: {all_sessions}
Sessions drilled (key facts read): {drilled}
Previous queries tried: {prev_queries}

Key facts gathered:
{gathered}

Your task:
1. ASSESS: Are facts sufficient for a COMPLETE
   answer?
2. CONVERGENCE CHECK: If same sessions keep
   appearing and no new ones found, mark
   sufficient.
3. ENTITY EXPANSION: Extract entities from
   facts as potential new search terms.
4. If NOT sufficient, generate 1-5 NEW search
   queries (3-8 words, different vocabulary
   from all previous queries).

Return JSON:
{"is_sufficient": true/false,
 "items_found_so_far": [...],
 "potential_gaps": "...",
 "new_queries": [...]}
\end{lstlisting}
\end{minipage}
\caption{Prompt for the \texttt{reflect} tool
(\S\ref{sec:agent}).}
\label{fig:prompt-reflect}
\end{figure}

\paragraph{Evidence organization and answer generation.}
The organized notes synthesizer prompt (Figure~\ref{fig:prompt-on-text}) instructs the model to structure retrieved evidence by relevance, temporal order, contradictions, and enumeration without producing a final answer. The answer generator (Figure~\ref{fig:prompt-answer}) then independently verifies the evidence and produces the final response; this prompt is shared with SimpleMem for fair comparison.

\begin{figure*}[t]
\centering
\begin{minipage}{\textwidth}
\begin{lstlisting}[style=promptstyle]
Read the question and all the retrieved evidence below. Produce organized notes that will help
an answer generator answer the question correctly.

## Question (asked on {question_date})
{question}

## Retrieved Evidence
{evidence_text}

## Instructions
Produce your notes following this structure:

1. **RELEVANT FACTS**: List every fact from the evidence that is directly relevant to answering
   the question. For each fact, note which session it comes from and the session date.
   Order by relevance (most relevant first).

2. **TEMPORAL ORDERING** (if dates matter): Arrange relevant facts chronologically. Clearly mark
   which is the MOST RECENT. If the question asks about the current/latest state, highlight it.

3. **CONTRADICTIONS / UPDATES**: If multiple sessions contain conflicting information about
   the same topic, note ALL versions with their dates. The LATEST date = current truth.

4. **ENUMERATION** (if "how many" or "list all"): Enumerate every qualifying item found.
   Deduplicate items appearing in multiple sessions. Flag uncertain items.

5. **ANSWER HINT**: Based on your organization of the evidence, what answer does the
   evidence point to? (This is a hint, not a final answer.)

Rules: Include EVERY relevant detail -- do not summarize away specifics (names, numbers,
dates, titles, locations). Be exhaustive for counting questions.
\end{lstlisting}
\end{minipage}
\caption{Prompt for Organized Notes Synthesizer (\S\ref{sec:organized-notes}). The system
message instructs the model to organize evidence without answering the question itself.}
\label{fig:prompt-on-text}
\end{figure*}

\begin{figure}[t]
\centering
\begin{minipage}{\columnwidth}
\begin{lstlisting}[style=promptstyle]
System: You are a professional Q&A assistant.
Extract concise answers from context.
You must output valid JSON format.

User:
Answer the user's question based on the
provided context.

User Question (asked on {question_date}):
{query}

Relevant Context:
{context_str}

Requirements:
1. Think through the reasoning process.
2. Provide a CONCISE but COMPLETE answer --
   include specific names, titles, key details.
3. Answer must be based ONLY on context.
4. Independently scan ALL context entries.
   Do NOT simply trust the Agent's preliminary
   answer -- verify it yourself.
5. Return JSON format.

Output Format:
{"reasoning": "Brief thought process",
 "answer": "Concise answer in a short phrase"}
\end{lstlisting}
\end{minipage}
\caption{Answer generator prompt (\S\ref{sec:organized-notes}).
Identical to SimpleMem's prompt for fair comparison; only the
input context differs.}
\label{fig:prompt-answer}
\end{figure}

\section{Hyperparameter Configuration}
\label{app:hyperparams}

Table~\ref{tab:hyperparams} lists all hyperparameters used in the
APDMem experiments. Memory construction parameters ($L_0$
window size and $L_1$ bullet bounds) are fixed across all
backbone models. Retrieval parameters were selected based on
preliminary experiments on a held-out development set of 20
questions and kept constant for all reported results.

\begin{table}[t]
\centering
\small
\begin{tabularx}{\columnwidth}{@{}lXl@{}}
\toprule
\textbf{Module} & \textbf{Parameter} & \textbf{Value} \\
\midrule
$L_0$ & Turn window size $W$ & 4 turns \\
$L_0$ & Window overlap & 1 turn \\
$L_0$ & Max chars/window & 6000 \\
\midrule
$L_1$ & Max input chars & 12000 \\
$L_1$ & Bullet count & 5--25 \\
$L_1$ & Generation & On-demand \\
$L_1$ & Workers & 8 \\
\midrule
Embedding & Model & Nomic-v1.5 \\
Embedding & Dimension & 768 \\
\midrule
Retrieval & RRF constant $k$ & 60 \\
Retrieval & Cosine weight $w_{\cos}$ & 0.5 \\
Retrieval & Top-$K$ sessions & 20 \\
Retrieval & Keyword boost & 0.1 \\
\midrule
Agent & Max ReAct steps & 20 \\
Agent & Auto-drill/search & 3 \\
Agent & Max drill sessions & 10 \\
Agent & Max raw turns & 20 \\
Agent & Max reflection rounds & 2 \\
\midrule
LLM & Agent temperature & 0.0 \\
\bottomrule
\end{tabularx}
\caption{Hyperparameter configuration for APDMem.}
\label{tab:hyperparams}
\end{table}

\section{Setting the Controller Step Budget}
\label{app:tmax}

The controller step budget $T_{\max}$ is an explicit inference-time
budget parameter. It caps the maximum number of convergence-control
iterations and therefore bounds the maximum number of additional
tool/LLM calls the controller can make before it must stop escalating
and return the best available answer.

\paragraph{Experimental setting.}
In all experiments, we set $T_{\max}=20$. This is intentionally a
generous cap: when the controller converges earlier, it terminates
before reaching the budget, while $T_{\max}$ prevents rare difficult
queries from incurring unbounded cost or latency. Table~\ref{tab:tmax}
reports the observed iteration distribution under this setting.

\begin{table}[t]
\centering
\small
\begin{tabular}{@{}lrrrr@{}}
\toprule
\textbf{Model} & \textbf{Median} & \textbf{p95} & \textbf{p99} & \textbf{Max} \\
\midrule
GPT-4.1      & 5 & 12 & 17 & 20 \\
GPT-4.1-mini & 3 & 12 & 16 & 19 \\
\bottomrule
\end{tabular}
\caption{Observed controller iteration statistics with $T_{\max}=20$.
The cap is reached only in rare cases, so it primarily acts as a
worst-case cost and latency bound.}
\label{tab:tmax}
\end{table}

With $T_{\max}=20$, nearly all queries terminate before reaching the
cap. Thus, in our experiments, the budget does not materially affect
answer quality for most examples, but it still provides an explicit
worst-case compute bound.

\paragraph{Choosing $T_{\max}$ in other applications.}
$T_{\max}$ controls a quality--cost trade-off: larger values allow the
controller to continue searching and reflecting on difficult queries,
whereas smaller values reduce worst-case latency and tool/LLM usage. In
practice, we recommend setting $T_{\max}$ as follows:

\begin{enumerate}
    \item \textbf{Profile demand.} On a representative sample, measure
    the empirical iteration distribution and set $T_{\max}$ near a high
    percentile, such as p95--p99, so truncation is rare.
    \item \textbf{Apply operational bounds.} Ensure the resulting
    maximum stays within deployment constraints, including latency and
    cost budgets.
    \item \textbf{Choose the tightest sufficient bound.} Use the
    smallest $T_{\max}$ that keeps truncation rare while satisfying the
    operational constraints.
\end{enumerate}

\section{Statistical Significance Analysis}
\label{app:stat-sig}

To further validate the robustness of the main results, we conduct a controlled multi-run evaluation comparing APDMem with SimpleMem, the strongest prior baseline. Both methods are evaluated on the full LongMemEval-S benchmark, which contains 500 questions, using GPT-4.1 as the backbone model. We run each method for 10 independent runs under the same evaluation protocol and report the mean and standard deviation of accuracy across runs.

\begin{table}[t]
\centering
\small
\begin{tabular}{l c}
\toprule
\textbf{Method} & \textbf{Accuracy (\%)} \\
\midrule
APDMem & $87.5 \pm 1.0$ \\
SimpleMem & $84.1 \pm 0.7$ \\
\bottomrule
\end{tabular}
\caption{
Multi-run accuracy on the full LongMemEval-S benchmark using GPT-4.1. We report mean and standard deviation over 10 independent runs. APDMem improves over SimpleMem by +3.4 percentage points on average.
}
\label{tab:stat-sig-accuracy}
\end{table}

APDMem achieves an average accuracy of 87.5\%, compared with 84.1\% for SimpleMem, yielding an average improvement of +3.4 percentage points. To test whether this improvement is statistically significant, we report both a parametric significance test and a non-parametric confidence interval.

As the primary parametric test, we use Welch's t-test, which compares the run-level means while allowing the two methods to have unequal variances. As a complementary robustness check, we also compute a bias-corrected and accelerated bootstrap confidence interval, which makes fewer distributional assumptions. The results are shown in Table~\ref{tab:stat-sig-tests}.

\begin{table*}[t]
\centering
\small
\begin{tabular}{l c c c}
\toprule
\textbf{Test} & \textbf{Type} & \textbf{95\% CI of $\Delta$} & \textbf{Statistic / Details} \\
\midrule
Welch's t-test & Parametric & $[+2.55, +4.25]$ & $t = 8.44$, $p = 2.6 \times 10^{-7}$ \\
BCa bootstrap & Non-parametric & $[+2.70, +4.20]$ & $B = 10{,}000$ \\
\bottomrule
\end{tabular}
\caption{
Statistical significance analysis of the APDMem--SimpleMem accuracy gap on LongMemEval-S using GPT-4.1. Here, $\Delta$ denotes the accuracy difference between APDMem and SimpleMem in percentage points. Both confidence intervals are strictly above zero, indicating that the improvement is statistically significant.
}
\label{tab:stat-sig-tests}
\end{table*}

Both confidence intervals are well above zero and provide consistent estimates of the APDMem--SimpleMem performance gap. These results indicate that, when evaluated using GPT-4.1, the observed improvement is statistically significant and unlikely to be explained by run-to-run variation alone.

\section{Human Validation of the LLM-as-a-Judge}
\label{app:judge-validation}

All accuracy results in our main experiments use the LongMemEval
LLM-as-a-judge protocol. Following prior work, we use the official judge prompt
with GPT-4.1-mini at temperature $0.0$, applied uniformly to APDMem and all
baselines. Since our reported gains depend on this evaluation protocol, we
conducted an additional human validation study to verify that the judge is
reliable in our setting.

\paragraph{Human verification.}
We manually validated a stratified subset of 120 question--prediction pairs,
with 20 examples from each of the six LongMemEval-S categories. One author,
blind to the judge decision, labeled each prediction as \textsc{Correct} or
\textsc{Wrong} according to whether it matched the gold answer, using the same
criterion as the judge. To stress-test reliability, we over-sampled examples
that the judge labeled \textsc{Wrong}, enriching the subset with borderline
cases where disagreements are more likely.

\paragraph{Agreement.}
Human and judge labels agreed on 119 of 120 examples, corresponding to 99.2\%
agreement and Cohen's $\kappa=0.98$, indicating near-perfect agreement.
Agreement was consistently high across all categories, as shown in
Table~\ref{tab:judge-validation}.

\begin{table}[t]
\centering
\small
\setlength{\tabcolsep}{10pt}
\begin{tabular}{@{}lccc@{}}
\toprule
Category & $n$ & Agree. & $\kappa$ \\
\midrule
knowledge-update & 20 & 0.950 & 0.886 \\
multi-session & 20 & 1.000 & 1.000 \\
single-session-assistant & 20 & 1.000 & 1.000 \\
single-session-preference & 20 & 1.000 & 1.000 \\
single-session-user & 20 & 1.000 & 1.000 \\
temporal-reasoning & 20 & 1.000 & 1.000 \\
\midrule
Overall & 120 & 0.992 & 0.980 \\
\bottomrule
\end{tabular}
\caption{Human validation of the LongMemEval judge on 120 stratified
question--prediction pairs.}
\label{tab:judge-validation}
\end{table}

The single disagreement was a case where the human annotator marked the
prediction as correct while the judge marked it as wrong, suggesting that the
judge is not biased in favor of APDMem. Overall, this validation provides
direct evidence that the LongMemEval judge is reliable in our setting and
mitigates the concern that the reported gains are driven by an unvalidated
LLM-as-a-judge protocol.

\section{Hybrid Retriever}
\label{app:retrieval}

The Hybrid Retriever retrieves candidate sessions using two
complementary indexes over all $L_0$ thematic-summary units
$\{c_t\}$.

\paragraph{Lexical index.}
A BM25 inverted index with stemming indexes each $c_t$ and its
keyword set $\mathcal{K}_t$ as separate fields, capturing exact
matches for entities, proper nouns, actions, and dates.

\paragraph{Dense index.}
Each $c_t$ is embedded with a sentence encoder
$f_{\text{enc}}$ using task-specific prefixes. Embeddings are
L2-normalized, reducing dense retrieval to dot-product search.

\paragraph{Reciprocal Rank Fusion.}
Given a query $q$, the lexical and dense indexes each produce a
ranking over turns. The Hybrid Retriever combines them using
Reciprocal Rank Fusion (RRF) \cite{cormack2009reciprocal}:
\begin{equation}
  \text{rrf}(q, c_t) =
    \sum_{j \in \{\text{lex}, \text{dense}\}}
    \frac{w_j}{k + r_j(c_t)},
  \label{eq:rrf}
\end{equation}
where $r_j(c_t)$ is the 1-based rank from index $j$, $k$ is a
smoothing constant, and $w_j$ are equal weights.

\paragraph{Session scoring.}
Each session is scored by its best-matching turn:
\begin{equation}
  s(q, S_i) = \max_{t \in S_i}\; \text{rrf}(q, c_t).
  \label{eq:sess_score}
\end{equation}
To preserve lexical precision for entity-heavy queries, the
Hybrid Retriever adds a keyword-field boost for exact matches:
\begin{equation}
  s'(q, S_i) = s(q, S_i) + \alpha \cdot s_{\max} \cdot n_i,
  \label{eq:boost}
\end{equation}
where $s_{\max}$ is the top hybrid score, $n_i$ is a discretized
match count, and $\alpha$ is a small constant.

\paragraph{Multi-query fusion.}
For decomposed questions, the controller may issue multiple
queries $Q = \{q_1,\ldots,q_m\}$. The Hybrid Retriever scores
each session by its best match across the query set:
\begin{equation}
  s_{\text{final}}(S_i) = \max_{q_j \in Q}\; s'(q_j, S_i).
  \label{eq:multiquery}
\end{equation}
The top-$K$ sessions are returned with their thematic summaries.

\section{Per-Model Efficiency Breakdown}
\label{app:efficiency-models}

Table~\ref{tab:efficiency-models} reports the per-model
token usage and latency for both offline memory construction
and online retrieval. All numbers are per-question averages
on LongMemEval ($n{=}500$).

\begin{table}[t]
\centering
\footnotesize
\setlength{\tabcolsep}{3.5pt}
\renewcommand{\arraystretch}{1.05}
\resizebox{\columnwidth}{!}{%
\begin{tabular}{@{}l rrr rrr r@{}}
\toprule
& \multicolumn{3}{c}{\textbf{Offline (Construction)}}
  & \multicolumn{4}{c}{\textbf{Online (Controller)}} \\
\cmidrule(lr){2-4} \cmidrule(lr){5-8}
\textbf{Model}
  & \textbf{In} & \textbf{Out} & \textbf{Tot.}
  & \textbf{In} & \textbf{Out} & \textbf{Tot.}
  & \textbf{Lat.} \\
\midrule
GPT-4o-mini   & 130.1 & 15.2 & 145.4
  & 121.5 & 2.8 & 124.3 & 37.7s \\
GPT-4.1-mini  & 130.1 & 20.6 & 150.7
  & 86.9  & 2.4 & 89.3  & 57.7s \\
GPT-4.1       & 130.1 & 21.2 & 151.3
  & 125.3 & 3.2 & 128.5 & 45.9s \\
GPT-5.2       & 130.1 & 23.7 & 153.8
  & 84.8  & 1.7 & 86.5  & 42.7s \\
\bottomrule
\end{tabular}}
\caption{Per-model efficiency breakdown (per-question
averages, LongMemEval $n{=}500$). Input/output token counts
and total are in thousands (k). Offline cost is identical
across models except for output tokens; online cost reflects
the agentic controller only.}
\label{tab:efficiency-models}
\end{table}

\paragraph{Offline cost.}
All models share the same input cost (130.1k tokens) because
$L_0$ generation processes identical conversation windows.
Output tokens increase slightly with model capability
(15.2k for GPT-4o-mini vs.\ 23.7k for GPT-5.2), reflecting
richer summaries and keywords, but the total remains within
a narrow 145--154k range.

\paragraph{Online cost.}
Stronger models exhibit more efficient retrieval:
GPT-4.1-mini and GPT-5.2 consume only 86--89k tokens
compared to 121--129k for GPT-4o-mini and GPT-4.1.
This suggests that more capable models require fewer ReAct
steps to converge, they generate better initial search
queries and identify evidence gaps earlier, reducing
iterative exploration. GPT-5.2 achieves the lowest token
cost (86.5k) \emph{and} competitive latency (42.7s),
indicating that its retrieval efficiency translates to both
cost and speed improvements. GPT-4.1-mini shows the highest
latency (57.7s) despite low token usage, likely due to
slower per-request API response times.

\section{Multi-Resolution Memory Substrate: Construction and Examples}
\label{app:memory-details}

This section describes the construction process for each layer of
the Multi-Resolution Memory Substrate (\S\ref{sec:memory}) and
provides real output examples from the LongMemEval dataset.
Prompts for $L_0$, $L_1$, and $L_2$ are shown in
Figures~\ref{fig:prompt-level0},~\ref{fig:prompt-level1},
and~\ref{fig:prompt-level2}, respectively. All examples below
are from the same user (\texttt{0a995998}, 44~sessions).

\subsection{$L_0$: Thematic Summaries}
\label{app:detail-level0}

\paragraph{Construction.}
Each session is split into user--assistant turns.
A sliding window of $W{=}4$ turns (overlap 1) is passed to the
LLM, which produces a one-sentence summary (max 30~words),
5--8 keywords, and 3--5 tags per turn. Pronouns are resolved and
relative dates are converted to absolute dates using the session
timestamp as anchor. Summaries are stored in Markdown and
embedded with a sentence encoder for dense retrieval.
$L_0$ is materialized at ingestion time. Figure~\ref{fig:output-level0} shows a real $L_0$ output.

\begin{figure}[t]
\centering
\begin{minipage}{\columnwidth}
\begin{lstlisting}[style=promptstyle]
## Session 0 (date=2023/02/15, id=f10be626)

Turn 1:
- Summary: 2023-02-15 conversation covered
  solo-trip destination ideas for user with
  $2000 budget over 5-7 days.
- Keywords: 2023-02-15, solo trip planning,
  $2000 budget, 5-7 days, Airbnb, Southeast
  Asia, Portugal, Costa Rica
- Tags: travel planning, budgeting,
  accommodations, solo travel

Turn 2:
- Summary: Assistant recommended Lisbon and
  Porto in Portugal with estimated daily
  costs of $80-120.
- Keywords: Lisbon, Porto, Portugal, daily
  budget $80-120, hostels, local food
- Tags: destination recommendation, Europe,
  budget travel
\end{lstlisting}
\end{minipage}
\caption{$L_0$ output example: per-turn thematic summaries
with keywords and tags.}
\label{fig:output-level0}
\end{figure}

\subsection{$L_1$: Personalized Key Facts}
\label{app:detail-level1}

\paragraph{Construction.}
When the progressive disclosure controller calls
\texttt{read\_facts} on a session, the LLM extracts a bullet
list of 5--25 self-contained key facts about user information,
preferences, plans, and constraints. Each bullet resolves all
pronouns and relative dates. Generated facts are cached for
subsequent accesses. $L_1$ is generated on demand (lazy
materialization). Figure~\ref{fig:output-level1} shows a real $L_1$ output.

\begin{figure}[t]
\centering
\begin{minipage}{\columnwidth}
\begin{lstlisting}[style=promptstyle]
## Session 2 (date=2023/02/15, id=750b437b_3)

- User was looking for recipe ideas that
  incorporate seasonal ingredients.
- User made a batch of roasted root
  vegetables before 2023-02-15.
- User was interested in citrus fruit
  recipes, including oranges and lemons.
- User stated that oranges and lemons are
  in season from 2022-12 to 2023-05.
- User was particularly interested in
  trying Citrus and Avocado Salad.
- User was particularly interested in
  trying Lemon and Herb Roasted Chicken.
- User planned to try Citrus and Avocado
  Salad for a lunch break at work.
- User usually took a 10-minute walk
  outside during a work break.
- User spotted daffodils blooming in a
  nearby park before 2023-02-15.
- User had been experimenting with spring
  desserts including lemon bars.
\end{lstlisting}
\end{minipage}
\caption{$L_1$ output example: self-contained key facts
extracted from a single session.}
\label{fig:output-level1}
\end{figure}

\subsection{$L_2$: Turn-Level Evidence Notes}
\label{app:detail-level2}

\paragraph{Construction.}
When the controller calls \texttt{read\_turns} on a session,
the LLM decomposes each turn into multiple atomic fact bullets
grouped by turn number. Each bullet is a complete, independent,
self-contained statement preserving all specifics (names, dates,
quantities). Like $L_1$, $L_2$ is generated on demand and
cached. The prompt (Figure~\ref{fig:prompt-level2}) instructs
the model to split distinct facts into separate bullets and
never merge multiple facts into one. Figure~\ref{fig:output-level2} shows a real $L_2$ output.

\begin{figure}[t]
\centering
\begin{minipage}{\columnwidth}
\begin{lstlisting}[style=promptstyle]
## Session 7 (date=2023/05/21)

Turn 1:
- The user asked how hydraulic fracturing
  affected groundwater quality in the
  Marcellus Shale region.
- The assistant stated that the Marcellus
  Shale region was located in the eastern
  United States.
- The assistant stated that some studies
  found higher levels of methane, benzene,
  and arsenic in groundwater near hydraulic
  fracturing wells.
- The assistant stated that other studies
  found no significant groundwater quality
  changes near hydraulic fracturing wells.
- The assistant stated that Pennsylvania
  required hydraulic fracturing companies
  to monitor groundwater quality at nearby
  wells before drilling.

Turn 2:
- The user stated distrust of hydraulic
  fracturing companies properly monitoring
  operations.
- The assistant stated that the EPA set and
  enforced regulations on hydraulic
  fracturing operations.
- The assistant stated that citizens could
  report suspected violations or file
  lawsuits against hydraulic fracturing
  companies.
\end{lstlisting}
\end{minipage}
\caption{$L_2$ output example: atomic fact bullets grouped
by turn number.}
\label{fig:output-level2}
\end{figure}

\subsection{$L_3$: Raw Messages}
\label{app:detail-level3}

\paragraph{Construction.}
No LLM processing is required. The verbatim user--assistant
messages are stored directly, grouped by turn number. $L_3$
is materialized at ingestion time by simply formatting the
original conversation into a structured Markdown file. The
controller accesses $L_3$ via \texttt{read\_raw} when exact
wording or fine-grained details not preserved in higher layers
are needed. Figure~\ref{fig:output-level3} shows a real $L_3$ output.

\begin{figure}[t]
\centering
\begin{minipage}{\columnwidth}
\begin{lstlisting}[style=promptstyle]
## Session 0 (date=2023/02/15, id=f10be626)

### Turn 1

User:
I'm trying to plan my next trip and was
thinking about my past travel experiences.
Can you help me find some good destinations
for a solo trip?

Assistant:
I'd be happy to help you plan your next
solo adventure. To get started, can you
tell me:
1. What type of activities do you enjoy?
2. What's your budget for the trip?
3. How long do you have for the trip?
4. Are you open to exploring new destinations?
5. What type of accommodation do you prefer?
\end{lstlisting}
\end{minipage}
\caption{$L_3$ output example: verbatim user--assistant
messages grouped by turn.}
\label{fig:output-level3}
\end{figure}

\section{Generator-Side Ablation}
\label{app:ablation-gen}

To isolate the generator's contribution, we fix the retrieval
results from a single run and vary only the generation
strategy: whether evidence passes through the \textit{Organized
Notes Synthesizer} or is fed raw to the answer generator, and
whether the controller's draft answer is included.
Table~\ref{tab:ablation-gen} reports results.

\begin{table}[t]
\centering
\small
\begin{tabular}{@{}lc@{}}
\toprule
\textbf{Configuration} & \textbf{Avg.\ Acc.\ (\%)} \\
\midrule
APDMem (full)                        & 87.8 \\
\quad w/o draft answer               & 86.6 \\
\quad w/o Organized Notes            & 85.2 \\
\quad w/o Organized Notes \& draft   & 83.6 \\
\bottomrule
\end{tabular}
\caption{Generator ablation: effect of Organized Notes and
draft answer on final accuracy (GPT-4.1, fixed retrieval).}
\label{tab:ablation-gen}
\end{table}

The full system achieves 87.8\%. Removing the draft answer
reduces accuracy by 1.2 points, while removing Organized
Notes causes a larger drop of 2.6 points. Without both,
accuracy falls to 83.6\% ($-4.2$), confirming that structured
evidence synthesis and the controller's draft reasoning
provide complementary benefits.

Organized Notes primarily helps categories requiring evidence
aggregation: \emph{multi-session synthesis} (+6.0 over raw)
and \emph{temporal reasoning} (+3.0), where the structured
format, which organizes facts by relevance, enumerates items
explicitly, and highlights contradictions, prevents the
generator from anchoring on a subset of evidence. The draft
answer primarily benefits single-session categories
(\emph{SS-User}: +2.9, \emph{SS-Assistant}: +1.8), where the
controller's focused reasoning about a specific session
provides useful context for the generator.

\end{document}